\documentclass{article}

\usepackage{fullpage} 

\usepackage[utf8]{inputenc} 
\usepackage[T1]{fontenc}    
\usepackage{hyperref}       
\usepackage{url}            
\usepackage{booktabs}       
\usepackage{amsfonts}       
\usepackage{nicefrac}       
\usepackage{microtype}      

\usepackage{amsmath, amsthm, amssymb, mathtools}
\usepackage{algorithm, algpseudocode} 

\usepackage[usenames]{xcolor}
\usepackage{bm, xspace}
\usepackage{pifont}
\usepackage{multirow, hhline, array, graphicx, multicol}
\usepackage{tikz, pgfplots}

\usepackage{natbib}
\usepackage{makecell}
\usepackage{enumitem}
\newtheorem{Definition}{Definition}
\newtheorem{Theorem}{Theorem}
\newtheorem{Lemma}{Lemma}
\newtheorem{Claim}{Claim}

\renewcommand{\cite}[1]{\citep{#1}}

\title{Autoencoders Learn Generative Linear Models}

\author{Thanh V. Nguyen, Raymond K. W. Wong, and Chinmay Hegde 
\thanks{Email: \{thanhng, chinmay\}@iastate.edu;
  raywong@tamu.edu. T. N. and C. H. are with the Electrical and
  Computer Engineering Department at Iowa State University. R. W. is with
  Statistics Department at Texas A\&M University. This work was
  supported in part by the National Science Foundation under grants
  CCF-1566281, CAREER CCF-1750920 and DMS-1612985, and in part by a Faculty Fellowship from the Black and Veatch Foundation.}
}

\begin{document}

\maketitle


\def\1{\bm{1}}

\newcommand*{\bigcdot}{\bullet}

\newcommand{\cA}[1]{A_{#1}} 
\newcommand{\cAi}{\cA{i}}
\newcommand{\cAj}{\cA{j}}
\newcommand{\cAl}{\cA{l}}
\newcommand{\cAS}{\cA{S}}

\newcommand{\cB}[1]{A_{#1}} 
\newcommand{\cBi}{\cB{i}}
\newcommand{\cBS}{\cB{S}}

\newcommand{\cD}[1]{D_{ #1}} 
\newcommand{\cDi}{\cD{i}}
\newcommand{\cDj}{\cD{j}}
\newcommand{\cDl}{\cD{l}}
\newcommand{\cDS}{\cD{S}}

\newcommand{\rA}[1]{A_{#1 \bigcdot}} 
\newcommand{\rAj}{\rA{j}}
\newcommand{\rAR}{\rA{R}}
\newcommand{\rAl}{\rA{l}}

\newcommand{\AR}[1]{A_{R, #1}}

\newcommand{\cg}[1]{g_{#1}} 
\newcommand{\cgi}{\cg{i}}
\newcommand{\cgS}{\cg{S}}
\newcommand{\gR}[1]{g_{R, #1}}

\def\ghat{\widehat{g}}
\def\Rhat{\widehat{R}}
\def\dhat{\widehat{d}}
\def\ehat{\widehat{e}}
\def\Muv{M_{u,v}}
\def\Mhuv{\widehat{M}_{u,v}}

\def\VarE{\mathcal{E}}
\def\Ne{N_{\varepsilon}}
\def\Rad{\mathcal{R}}
\def\Aupb{\mathcal{A}}
\def\Bupb{\mathcal{B}}
\def\di{\Delta_i}
\def\dj{\Delta_j}
\def\dl{\Delta_l}
\def\muij{\mu_{ij}}
\def\muil{\mu_{il}}
\def\mujl{\mu_{jl}}
\def\mux{\kappa_1}
\def\sigmax{\kappa_2}
\def\varx{\kappa}

\newcommand*{\Asmp}[1]{Assumption \textnormal{\textbf{A#1}}}
\newcommand*{\AsmpB}[1]{Assumption \textnormal{\textbf{B#1}}}

\def\sgn{\mathrm{sgn}}
\def\supp{\mathrm{supp}}
\def\diag{\mathrm{diag}}
\def\card{\textnormal{card}}
\def\thres{\mathrm{threshold}}
\def\nor{\mathrm{normalize}}
\def\polylog{\textnormal{polylog}}
\def\iid{\text{i.i.d.}}
\def\whp{\text{w.h.p.}}
\def\wrt{\text{w.r.t.}}
\def\ie{\text{i.e.}}
\def\eg{\text{e.g.}}
\def\relu{\mathrm{ReLU}}

\newcommand{\ceil}[1]{\left \lceil #1 \right \rceil}

\def\cmark{\ding{51}}
\def\xmark{\ding{55}}

\newcommand{\E}{\mathbb{E}}
\newcommand{\var}{\mathrm{Var}}
\newcommand{\Prob}{\mathbb{P}}
\newcommand{\sigmae}{\sigma_\eta}
\newcommand{\HardThres}{\mathcal{H}}

\newcommand{\R}{\mathbb{R}}

\DeclarePairedDelimiterX{\norm}[1]{\lVert}{\rVert}{#1} 
\DeclarePairedDelimiterX{\inprod}[2]{\langle}{\rangle}{#1, #2}
\DeclarePairedDelimiterX{\abs}[1]{\lvert}{\rvert}{#1}

\DeclarePairedDelimiterX{\bigO}[1]{(}{)}{#1}
\def\Otilde{\widetilde{O}}
\def\Omgtilde{\widetilde{\Omega}}

\newcommand{\sgnEvent}{\mathcal{F}_{x^*}}

\renewcommand{\qedsymbol}{$\blacksquare$}


\begin{abstract}
 
%
 We provide a series
of results for unsupervised learning with autoencoders. 
Specifically, we study shallow two-layer
autoencoder architectures with shared weights. We focus on three generative models for data that are common in statistical machine learning: (i) the mixture-of-gaussians model, (ii) the sparse coding
model, and (iii) the sparsity model with non-negative coefficients. 
For each of these models, we 
prove that under suitable
choices of hyperparameters, architectures, and initialization, 
autoencoders 
learned by gradient descent 
can successfully recover the parameters of the corresponding model. To
our knowledge, this is the first result that rigorously studies the
dynamics of gradient descent for weight-sharing autoencoders. 
Our analysis can be viewed as theoretical evidence that shallow autoencoder modules indeed can be used as feature learning mechanisms for a variety of data models, 
 and may shed insight on how to train larger stacked architectures with autoencoders as basic building blocks.
\end{abstract}

\section{Introduction}
\label{intro}

\subsection{Motivation}

Due to the resurgence of neural networks and deep learning, 
there
has been growing interest in the community towards a thorough and
principled understanding of 
training neural networks in both theoretical and algorithmic aspects. This has led to several important
breakthroughs recently, including provable algorithms for learning shallow ($1$-hidden layer) networks with nonlinear activations~\cite{tian2017symmetry,geleema18,brutkus-globerson,zhong}, deep networks with linear activations~\cite{kawaguchi}, and residual networks~\cite{li,hardtma}.

A typical approach adopted by this line of work is as follows: assume that the data obeys a ground truth \emph{generative model} (induced by simple but reasonably expressive data-generating distributions), and prove that the weights learned by the proposed algorithms (either exactly or approximately) recover the parameters of the generative model. Indeed, such distributional assumptions are necessary to overcome known NP-hardness barriers for learning neural networks~\cite{hardness}. Nevertheless, the majority of these approaches have focused on neural network architectures for supervised learning, barring a few exceptions which we detail below.

\subsection{Our contributions}

In this paper, we complement this line of work by providing new theoretical results for \emph{unsupervised} learning using neural networks. Our focus here is on shallow two-layer autoencoder architectures with 
shared weights. Conceptually, we build upon previous theoretical
results on learning autoencoder
networks~\cite{arora14_bound,arora15_reversible,tran17}, and we
elaborate on the novelty of our work in the discussion on prior work
below.

Our setting is standard: we assume that the training data consists of i.i.d.\ samples from a high-dimensional distribution parameterized by a generative model, and we train the weights of the autoencoder using ordinary (batch) gradient descent. We consider three families of generative models that are commonly adopted in machine learning: (i) the Gaussian mixture model with well-separated centers~\cite{arora05_mog}; (ii) the $k$-sparse model, specified by sparse linear combination of atoms~\cite{spielman12_exact}; and (iii) the non-negative $k$-sparse model~\cite{tran17}.
While these models are traditionally studied separately depending on the application, all of these model families can be expressed via a unified, generic form: 
\begin{equation}
\label{eq:lin_model}
y = Ax^* + \eta,
\end{equation}
which we (loosely) dub as the \emph{generative bilinear model}. In this form, $A$ is a groundtruth $n \times m$-matrix, $x^*$ is an $m$-dimensional latent code vector and $\eta$ is an independent $n$-dimensional random noise vector. Samples $y$'s are what we observe. Different choices of $n$ and $m$, as well as different assumptions on $A$ and $x^*$ lead to the  three aforementioned generative models.

Under these three generative models, and with suitable choice of hyper-parameters, initial estimates, and autoencoder architectures, we rigorously prove that:
 
\begin{quote}
  \emph{Two-layer autoencoders, trained with (normalized) gradient descent 
    over the reconstruction loss, provably learn the parameters of the underlying generative bilinear model.}
\end{quote}

To the best of our knowledge, our work is the first to analytically characterize the dynamics of gradient descent for training two-layer autoencoders. Our analysis can be viewed as theoretical evidence that shallow autoencoders can be used as feature learning mechanisms (provided the generative modeling assumptions hold), a view that seems to be widely adopted in practice. 
Our analysis highlights the following interesting conclusions:
(i) the activation function of the hidden (encoder) layer influences the choice of bias;
(ii) the bias of each hidden neuron in the encoder plays an important
role in achieving the convergence of the gradient descent; and (iii) the gradient dynamics depends on the complexity of the generative model. Further, we speculate that our analysis may shed insight on practical considerations for training deeper networks with stacked autoencoder layers as building blocks~\cite{arora14_bound}.

\subsection{Techniques}

Our analysis is built upon recent algorithmic developments in the sparse coding literature~\cite{agarwal14_learning,gribonval15_sample,arora15_neural}. Sparse coding corresponds to the setting where the synthesis coefficient vector $x^{*(i)}$ in \eqref{eq:lin_model} for each data sample $y^{(i)}$ is assumed to be $k$-sparse, i.e., $x^{*(i)}$ only has at most $k \ll m$ non-zero elements. The exact algorithms proposed in these papers are all quite different,
but at a high level, all these methods involve establishing a notion that we dub as ``support consistency''. Broadly speaking, for a given data sample $y^{(i)} = Ax^{*(i)} + \eta^{(i)}$, the idea is that when the parameter estimates are close to the ground truth, it is possible to accurately estimate the true support of the synthesis vector $x^{*(i)}$ for \emph{each} data sample $y^{(i)}$.

We extend this to a broader family of generative models to form a notion that we call {``code consistency''}. We prove that if initialized appropriately, the weights of the hidden (encoder) layer of the autoencoder provides useful information about the \emph{sign} pattern of the corresponding synthesis vectors for every data sample. Somewhat surprisingly, the choice of activation function of each neuron in the hidden layer plays an important role in establishing code consistency and affects the possible choices of bias.

The code consistency property is crucial for establishing the correctness of gradient descent over the reconstruction loss. This turns out to be rather tedious due to the weight sharing --- a complication which requires a substantial departure from the existing machinery for analysis of sparse coding algorithms --- and indeed forms the bulk of the technical difficulty in our proofs. Nevertheless, we are able to derive explicit \emph{linear} convergence rates for all the generative models listed above. We do not attempt to analyze other training schemes (such as stochastic gradient descent or dropout) but anticipate that our analysis may lead to further work along those directions.

\subsection{Comparison with prior work}

Recent advances in algorithmic learning theory has led to numerous provably efficient algorithms for learning Gaussian mixture models, sparse codes, topic models, and ICA (see~\cite{arora05_mog,moitra10_mog,arora12_ica,vempala14_fpca,spielman12_exact,agarwal14_learning,gribonval15_sample,arora15_neural} and references therein). We omit a complete treatment of prior work due to space constraints.

We would like to emphasize that we do \emph{not} propose a new algorithm or autoencoder architecture, nor are we the first to highlight the applicability of autoencoders with the aforementioned  generative models. Indeed, generative models such as $k$-sparsity models have served as the motivation for the development of deep stacked (denoising) autoencoders dating back to the work of~\cite{vincent2010stacked}. The paper~\cite{arora14_bound} proves that stacked weight-sharing autoencoders can recover the parameters of sparsity-based generative models, but their analysis succeeds only for certain generative models whose parameters are themselves randomly sampled from certain distributions. In contrast, our analysis holds for a broader class of networks; we make no randomness assumptions on the parameters of the generative models themselves. 

More recently, autoencoders have been shown to learn sparse representations~\cite{arpit15_ae}. The recent paper~\cite{tran17} demonstrates that under the sparse generative model, the standard squared-error reconstruction loss of ReLU autoencoders exhibits (with asymptotically many samples) critical points in a neighborhood of the ground truth dictionary. However, they do not analyze gradient dynamics, nor do they establish convergence rates. We complete this line of work by proving explicitly that gradient descent (with column-wise normalization) in the asymptotic limit exhibits linear convergence up to a radius around the ground truth parameters.




\section{Preliminaries}
\paragraph{Notation}
Denote by $x_S$ the sub-vector of $x\in\R^m$
indexed by the elements of $S\subseteq [m]$. Similarly, let
$W_S$ be the sub-matrix of $W\in\R^{n\times m}$ with columns
indexed by elements in $S$
.
Also, define $\supp(x)\triangleq \{i\in[m]:x_i\neq 0\}$ as
the support of $x$
, $\sgn(x)$ as the element-wise sign of $x$ and $\1_E$ as the
indicator of an event $E$.

We adopt standard asymptotic notations: let $f(n) = O(g(n))$ (or $f(n) = \Omega(g(n))$) if there exists some constant $C >
0$ such that $|f(n)| \le C|g(n)|$ (respectively, $|f(n)| \ge
C|g(n)|$). Next, $f(n) = \Theta(g(n))$ is equivalent to that $f(n) =
O(g(n))$ and $f(n) = \Omega(g(n))$.
Also, $f(n)=\omega(g(n))$ if $\lim_{n\rightarrow\infty}|f(n)/g(n)|=\infty$.
In addition, $g(n)=O^*(f(n))$ indicates $|g(n)|\le K|f(n)|$
for some small enough constant $K$. Throughout, we use the phrase ``with high probability'' (abbreviated to \whp) to describe any event with failure probability at most $n^{-\omega(1)}$.

\subsection{Two-Layer Autoencoders}
\label{sec:autoencoder}

We focus on {shallow} autoencoders with a single hidden layer, $n$ neurons in the input/output layer and $m$ hidden neurons. We consider the \emph{weight-sharing} architecture in which the encoder has weights $W^T \in \R^{m\times n}$ and the decoder uses the shared weight $W \in \R^{n\times m}$. 
Denote $b \in \R^m$ as the vector of biases for the encoder (we do not consider decoder bias.) As such, for a given data sample $y \in \R^n$, the encoding and decoding respectively can be modeled as:
\begin{equation}
  x = \sigma(W^Ty + b)\quad\text{and}\quad\hat{y} = Wx,
  \label{eq:auto}
\end{equation}
where $\sigma(\cdot)$ denotes
the activation function in the encoder neurons. We consider two types of
activation functions: (i) the rectified linear unit: $$\relu(z) = \max(z,
0),$$ and (ii) the hard
thresholding operator: $$\mathrm{threshold}_{\lambda}(z) = z\1_{\abs{z} \ge \lambda}.$$ 
When applied to a vector (or matrix), these functions
are operated on each element and return a vector (respectively, matrix) of same size.
Our choice of the activation $\sigma(\cdot)$ function varies with different data generative models, and will be clear by context.


Herein, the loss function is the (squared) reconstruction error: 
\[
  L = \frac{1}{2} \norm{y - \hat{y}}^2 = \frac{1}{2} \norm{y -
    W\sigma(W^Ty + b)}^2,
\]
and we analyze the expected loss 
where the expectation is taken over the data distribution (specified below). 
Inspired by the literature of analysis of sparse
coding~\cite{agarwal13_exact,agarwal14_learning,tran17}, we
investigate the landscape of the expected loss so as to
shed light on dynamics of gradient descent for training the above autoencoder architectures. Indeed, we show
that for a variety of data distributions, such autoencoders can recover the distribution parameters
via suitably initialized gradient descent.

\subsection{Generative Bilinear Model}
\label{sec:gen_model}

We now describe an overarching generative model for the data samples. Specifically, we posit that the data samples $\{y^{(i)}\}^N_{i=1} \in \R^n$ are drawn according to the following ``bilinear'' model:
\begin{equation}
  \label{eq:gen_model}
  y = Ax^* + \eta,
\end{equation}
where $A \in \R^{n\times m}$  is a ground truth set of parameters, 
$x^* \in \R^m$ is a latent code vector, and $\eta\in \R^n$ represents noise.
Depending on different assumptions made on $A$ and $x^*$, this model generalizes various popular cases, such as mixture of spherical Gaussians, sparse coding, nonnegative sparse coding, and independent component analysis (ICA).
We will elaborate further on specific cases, but in general our generative model satisfies the following generic assumptions:
\begin{enumerate}
\item[A1.] The code $x^*$ is supported on set $S$ of size at most $k$, such that
  $p_i = \Prob[i \in S] = \Theta(k/m)$, $p_{ij} = \Prob[i, j \in S] = \Theta(k^2/m^2)$ and $p_{ijl} = \Prob[i, j, l \in S] = \Theta(k^3/m^3)$;
\item[A2.] Nonzero entries are independent; moreover, $\E[x_i^*|i\in S] = \kappa_1$ and $\E[x_i^{*2}|i\in S] =  \kappa_2 <\infty$;
\item[A3.] For $i \in S$, $\abs{x_i^*} \in [a_1, a_2]$ with $0 \le a_1 \leq a_2 \le \infty$;
\item[A4.] The noise term $\eta$ is distributed according to $\mathcal{N}(0, \sigmae^2I)$ and is independent of $x^*$.
\end{enumerate}


As special cases of the above model, we consider the following variants.

{\bf Mixture of spherical Gaussians:} 
We consider the standard Gaussian mixture model with $m$ centers, which is one of the most popular generative models encountered in machine learning applications. 
We model the means of the Gaussians as columns of the matrix $A$. To draw a data sample $y$, we 
sample $x^*$ uniformly from the canonical basis $\{e_i\}_{i=1}^m \in \R^n$ with probability $p_i=\Theta(1/m)$.
As such, $x^*$ has sparsity parameter $k=1$ with only one nonzero element being 1.
That means, $\kappa_1=\kappa_2=a_1=a_2=1$.


{\bf Sparse coding:} This is a well-known instance of the above structured linear
model, where the goal is basically to learn an \emph{overcomplete} dictionary $A$ that sparsely represents the input $y$. It has a rich history in various fields of signal processing, machine learning and
neuroscience~\cite{olshausen97_sc}. The generative model described above has successfully enabled recent theoretical advances in sparse coding~\citep{spielman12_exact,agarwal14_learning,gribonval15_sample,arora14_new_algorithms,arora15_neural}. The latent code vector $x^*$ is assumed to be $k$-sparse, whose nonzero entries are sub-Gaussian and bounded away from zero.
Therefore, $a_1>0$ and $a_2=\infty$. We assume that the distribution of nonzero entries are standardized such that $\kappa_1=0$, $\kappa_2=1$. Note that the condition of $\kappa_2$ further implies that $a_1\le 1$.

{\bf Non-negative sparse coding:} This is another variant of the above sparse coding model where the elements of the latent code $x^*$ are additionally required to be non-negative~\cite{tran17}. In some sense this is a generalization of the Gaussian mixture model described above. 
Since the code vector is non-negative, we \emph{do not} impose the
standardization as in the previous case of general sparse coding
($\kappa_1=0$ and $\kappa_2=1$); instead, we assume a compact interval
of the nonzero entries; that is, $a_1$ and $a_2$ are \emph{positive}
and \emph{bounded}. 



Having established probabilistic settings for these models, we now establish certain deterministic conditions on the true parameters $A$ to enable analysis. First, we require each column $A_i$ to be
normalized to unit norm in order to avoid the scaling ambiguity between $A$ and $x^*$. (Technically, this condition is not required for the mixture of Gaussian model case since $x^*$ is binary; however we make this assumption anyway to keep the treatment generic.) 
Second, we require columns of $A$ to be ``sufficiently distinct''; this is formalized by adopting the notion of pairwise incoherence. 

\begin{Definition}
  \label{def:incoherent}
  Suppose that $A \in \R^{n \times m}$ has unit-norm columns. $A$ is said to be $\mu$-incoherent if for every pair of column indices $(i,j),~i \neq j$ we have
$|\inprod{\cAi}{\cAj}| \leq \frac{\mu}{\sqrt{n}}$.
\end{Definition}

Though this definition is motivated from the sparse coding literature, pairwise incoherence is sufficiently general to enable identifiability of all aforementioned models. For the mixture of Gaussians with unit-norm means, pairwise incoherence states that the means are well-separated, which is a standard assumption. In the case of Gaussian mixtures, we assume that $m=O(1) \ll n$. For sparse coding, we focus on learning overcomplete dictionaries where $n \le m = O(n)$ . 
For the sparse coding case, we further require the spectral norm bound on $A$, \ie, $\norm{A} \le O(\sqrt{m/n})$. (In other words, $A$ is well-conditioned.)

Our eventual goal is to show that training autoencoder via gradient descent can effectively recover the generative model parameter $A$.
To this end, we need a measure of goodness in recovery. Noting that any recovery method can only recover $A$ up to a permutation ambiguity in the columns (and a sign-flip ambiguity in the case of sparse coding), we first define an operator $\pi$ that permutes the columns of the matrix (and multiplies by $+1$ or $-1$ individually to each column in the case of sparse coding.) Then, we define our measure of goodness:
\begin{Definition}[$\delta$-closeness and $(\delta, \xi)$-nearness]
  A matrix $W$ is said to be $\delta$-close to $A$ if there exists an operator $\pi(\cdot)$ defined above such that $\norm{\pi(W)_i - A_i} \leq \delta$ for all $i$.
  We say $W$ is $(\delta,\xi)$-near to $A$ if in addition $\norm{\pi(W) -
    A} \leq \xi \norm{A}$.
  \label{def:goodness}
\end{Definition}
To simplify notation, we simply replace $\pi$ by the identity operator while keeping in mind that we are only recovering an element from the equivalence class of all permutations and sign-flips of $A$.

Armed with the above definitions and assumptions, we are now ready to state our results. Since the actual mathematical guarantees are somewhat tedious and technical, we summarize our results in terms of informal theorem statements, and elaborate more precisely in the following sections.

Our first main result establishes the code consistency of weight-sharing autoencoders under all the generative linear models described above, provided that the weights are suitably initialized.

\begin{Theorem}[informal]
  Consider a sample $y = Ax^* + \eta$. Let $x = \sigma(W^Ty +b)$ be the output of the encoder part of the autoencoder. Suppose that
  $W$ is $\delta$-close to $A$ with $\delta = O^*(1/\log n)$.
  \begin{enumerate}[label=(\roman*),leftmargin=*]
  \item  If $\sigma(\cdot)$ is either the ReLU or the hard thresholding activation, then the support of the true code vector $x^*$ matches that of $x$ for the mixture-of-Gaussians and non-negative sparse coding generative models.
  \item If $\sigma(\cdot)$ is the hard thresholding activation, then the support of $x^*$ matches that of $x$ for the sparse coding generative model. 
  \end{enumerate}
  
\end{Theorem}

Our second main result leverages the above property. We show that iterative gradient descent over the weights $W$ \emph{linearly} converges to a small neighborhood of the ground truth.

\begin{Theorem}[informal]
  Provided that the initial weight $W^{0}$ such that $W^0$ is $(\delta, 2)$-near to $A$. Given
  asymptotically many samples drawn from the above models, an iterative gradient update of $W$ can linearly converge to a small neighborhood of the ground truth $A$.
\end{Theorem}

We formally present these technical results in the next sections. Note
that we analyze the encoding and the gradient given $W^s$ at iteration
$s$; however we often skip the superscript for clarity.


\section{Encoding Stage}
\label{sec:hidden-activation}

Our technical results start with the analysis of the encoding stage in the forward pass.
We rigorously prove that the encoding performed by the autoencoder is sufficiently good in the sense that it recovers part of the information in the latent code $x^*$ (specifically, the signed support of $x^*$.) This is achieved based on appropriate choices of activation function, biases, and a good $W$ within close neighborhood of the true parameters $A$. We call this property code consistency:

\begin{Theorem}[Code consistency]
  \label{thm:consistency}
  Let $x = \sigma(W^Ty +b)$. Suppose $W$ is $\delta$-close to $A$
  with $\delta = O^*(1/\log n)$ and the noise satisfies $\sigmae = O(1/\sqrt{n})$.
  Then the following results hold:
  \begin{enumerate}[label=(\roman*),leftmargin=*]

  \item General $k$-sparse code with thresholding activation:
    Suppose $\mu \leq \sqrt{n}/\log^2n$ and $k\leq n/\log n$.
    If $x = \thres_{\lambda}(W^Ty + b)$ with $\lambda = a_1/2$ and $b = 0$,
    then with high probability
    \[
      \sgn(x) = \sgn(x^*).
    \]
    
  \item Non-negative $k$-sparse code with ReLU activation:
    Suppose $\mu \leq \delta\sqrt{n}/k$ and $k = O(1/\delta^2)$.
    If $x = \relu(W^Ty +b)$, and $b_i \in [-(1 -
    \delta)a_1 + a_2\delta\sqrt{k},\, -a_2\delta\sqrt{k}]$ for all $i$,
    then with high probability,
    \[
      \supp(x) = \supp(x^*).
    \]

    \item Non-negative $k$-sparse code with thresholding activation:
      Suppose $\mu \leq \delta\sqrt{n}/k$ and $k = O(1/\delta^2)$.
      If $x = \thres_{\lambda}(W^Ty +b)$ with $\lambda = a_1/2$ and $b = 0$,
    then with high probability,
    \[
      \supp(x) = \supp(x^*).
    \]

\end{enumerate}
\end{Theorem}

The full proof for Theorem \ref{thm:consistency} is relegated to Appendix~\ref{sec:cst_proof}. Here, we provide a short proof for the mixture-of-Gaussians generative model, which is really a special case of $(ii)$ and $(iii)$ above, where $k = 1$ and the nonzero component of $x^*$ is equal to $1$ (\ie, $\mux = \sigmax = a_1 = a_2 = 1$.)

\proof Denote $z = W^Ty + b$ and $S=\supp(x^*) = \{j\}$. Let $i$ be
fixed and consider two cases: if $i=j$, then 
\[
  z_i = \inprod{W_i}{\cAi} + \inprod{W_i}{\eta}
    + b_i \geq (1 - \delta^2/2) - \sigmae\log n + b_i > 0 ,
\]
\whp\ due to the fact that $\inprod{W_i}{\cAi} \geq 1
- \delta^2/2$ (Claim~\ref{cl:bound_cross_prod}), and the conditions $\sigma_\eta=O(1/\sqrt{n})$ and $b_i > -1 + \delta$.

On the other hand, if $i \neq j$, then using Claims~\ref{cl:bound_cross_prod} and~\ref{cl:bound_noise} in Appendix~\ref{sec:cst_proof}, we have \whp
\[
  z_i = \inprod{W_i}{\cAj} + \inprod{W_i}{\eta}
    + b_i
  \leq \mu/\sqrt{n} + \delta + \sigmae\log n + b_i < 0 ,
\]
for $b_i \leq -2\delta, \mu \leq \delta\sqrt{n}/k$ and $\sigma_\eta= O(1/\sqrt{n})$.
Due to Claim 2, these results hold \whp\ uniformly for all $i$, and hence $x = \relu(z)$ has the same support as $x^*$ \whp .

Moreover, one can also see that when $b_i = 0$, then \whp, $z_i > 1/2$ if $i = j$ and $z_i < 1/4$ otherwise. This result holds \whp\ uniformly for all $i$, and therefore, $x = \thres_{1/2}(z)$ has the same support as $x^*$ \whp
\qedhere

Note that for the non-negative case, both $\relu$ and $\thres$
activation would
lead to a correct support of the code, but this requires  $k =
O(1/\delta^2)$, which is rather restrictive and might be a limitation of the current analysis. 
Also, in Theorem \ref{thm:consistency},
$b$ is required to be negative for ReLU activation
for any $\delta>0$ due to the error of the current estimate $W$.
However, this result is consistent with the conclusion of \cite{zero-bias}
that negative bias is desirable for ReLU activation to produce sparse code.
Note that such choices of $b$ also lead to statistical bias (error) in nonzero code
and make it difficult to construct a provably correct learning procedure (Section \ref{sec:learning}) for ReLU activation.

Part (i) of Theorem \ref{thm:consistency} mirrors the consistency result
established for sparse coding in~\cite{arora15_neural}. 

Next, we apply the above result to show that provided the consistency result a (batch) gradient
update of the weights $W$ (and bias in certain cases) converges to the
true model parameters.


\section{Learning Stage}
\label{sec:learning}

In this section, we show that a gradient descent update for $W$ of the autoencoder (followed by a normalization in the Euclidean column norm of the
updated weights) leads to a linear convergence to a small neighborhood
of the ground truth $A$ under the aforementioned generative
models. For this purpose, we analyze the gradient of the expected loss with respect to $W$. Our analysis involves calculating the expected value of the gradient as if we were given infinitely many samples. (The finite sample analysis is left as future work.)

Since both ReLU and hard thresholding activation functions are non-differentiable at some values,
we will formulate an approximate gradient. Whenever differentiable,
the gradient of the loss $L$ with respect to the \emph{column} $W_i \in \R^{n}$ of the weight matrix $W$ is given by:
\begin{equation}
  \label{eq:grad_general}
  \nabla_{W_i}L = -\sigma'(W_i^Ty + b_i)\bigl[(W_i^Ty + b_i)I +
  yW_i^T\bigr]\bigl[y - Wx \bigr],
\end{equation}
where $x = \sigma(W^Ty + b)$ and $\sigma'(z_i)$ is the gradient of $\sigma(z_i)$ at $z_i$ where $\sigma$ is differentiable.  For
the rectified linear unit $\relu(z_i) = \max(z_i, 0)$, its gradient is
\begin{equation*}
  \sigma'(z_i) =
  \begin{cases}
    1 \quad \text{if } z_i > 0 , \\
    0 \quad \text{if } z_i < 0 . 
  \end{cases}
\end{equation*}
On the other hand, for the hard thresholding activation $\thres_{\lambda}(z_i) = z_i\1_{\abs{z_i} \ge \lambda}$, the
gradient is
\begin{equation*}
  \sigma'(z_i) =
  \begin{cases}
    1 \quad \text{if } \abs{z_i} > \lambda , \\
    0 \quad \text{if } \abs{z_i} < \lambda .
  \end{cases}
\end{equation*}
One can see that in both cases, the gradient $\sigma'(\cdot)$ at $z_i = W_i^Ty +
b_i$ resembles an indicator function $\1_{x_i \neq
  0}=\1_{\sigma(z_i)\neq 0}$ except where it is not defined. The observation motivates us to approximate the
$\nabla_{W_i}L$ with a simpler rule by replacing $\sigma'(W_i^Ty + b_i)$
with $\1_{x_i \neq 0}$:
\[
  \widetilde{\nabla_iL} = -\1_{x_i \neq 0}(W_i^TyI + b_iI+ yW_i^T)(y - W x).
\]
In fact,~\cite{tran17} (Lemma 5.1) shows that this approximate gradient $\widetilde{\nabla_iL}$ is a good
approximation of the true gradient~\eqref{eq:grad_general} in expectation.
Since $A$ is assumed to have normalized columns (with $\|A_i\|=1$),
we can enforce this property to the update by a simple column normalization after every update;
to denote this, we use the operator $\nor(\cdot)$ that returns a matrix $\nor(B)$ with unit columns, i.e.: 
$$\nor(B)_i=B_i/\|B_i\| ,$$ 
for any matrix $B$ that has no all-zero columns. 

Our convergence result leverages the code consistency property in Theorem \ref{thm:consistency}, but in turn succeeds under constraints on the biases of the hidden neurons $b$.
For thresholding activation, we can show that the simple choice of setting all biases to zero leads to both code consistency and linear convergence.
However, for ReLU activation, the range of bias specified in
Theorem~\ref{thm:consistency} (ii) has a profound effect on the descent procedure.
Roughly speaking, we need non-zero bias in order to ensure code consistency, but high values of bias can adversely impact gradient descent.
Indeed, our current analysis does not succeed for any \emph{constant} choice of bias (i.e., we do not find a constant bias that leads to both support consistency and linear convergence.)
To resolve this issue, we propose to use a simple \emph{diminishing} (in magnitude) sequence of biases $b$ along different iterations of the algorithm. Overall, this combination of approximate gradient and normalization lead to an update rule that certifies the existence of a linear convergent algorithm (up to a neighborhood of $A$.) 
The results are formally stated as follows:

\begin{Theorem}[Descent property]
  \label{thm:descent}
  Suppose that at step $s$ the weight $W^s$ is $(\delta^s, 2)$-near to
  $A$. There exists an iterative update rule using an approximate gradient $g^s$: $W^{s+1} = \nor(W^{s}
  -\zeta g^s)$ that linearly converges to $A$ when given infinitely many \emph{fresh} samples. More precisely, there exists some $\tau \in (1/2, 1)$ such that:
  \begin{enumerate}[label=(\roman*),leftmargin=*]
  \item Mixture of Gaussians: 
    Suppose the conditions in either (ii) or (iii) of Theorem~\ref{thm:consistency} hold.
    Suppose that the learning rate $\zeta = \Theta(m)$, and that the bias vector $b$ satisfies:
    \begin{itemize}
    \item[(i.1)] $b = 0$ if $x = \thres_{1/2}(W^Ty + b)$; or
    \item[(i.2)] $b^{s+1} = b^s/C$ if $x = \relu(W^Ty + b)$ for some constant $C > 1$. 
    \end{itemize}
    Then,
      $\norm{W^{s+1} - A}_F^2 \leq (1-\tau)\norm{W^{s} - A}_F^2 + O(mn^{-O(1)}).$

  \item General $k$-sparse code:
    Provided the conditions in Theorem~\ref{thm:consistency} (i) hold
    and the learning rate $\zeta = \Theta(m/k)$.
   
   Then,
     $ \norm{W^{s+1} - A}_F^2 \leq (1-\tau)\norm{W^{s} - A}_F^2 + O(mk^2/n^2).$

  \item Non-negative $k$-sparse code:
    Suppose the conditions in either (ii) or (iii) of
    Theorem~\ref{thm:consistency} hold. Suppose that the learning rate
    $\zeta = \Theta(m/k)$ and the bias $b$ satisfies:
    \begin{itemize}
    \item[(iii.1)] $b = 0$ if $x = \thres_{a_1/2}(W^Ty + b)$; or
    \item[(iii.2)] $b^{s+1} =  b^s/C$ if $x = \relu(W^Ty + b)$  for some constant $C > 1$. 
    \end{itemize}
    Then, 
    $  \norm{W^{s+1} - A}_F^2 \leq (1-\tau)\norm{W^{s} - A}_F^2 + O(k^3/m). $
  \end{enumerate}
\end{Theorem}

Recall the approximate gradient of the squared loss:
\[
  \widetilde{\nabla_iL} = -\1_{x_i \neq 0}(W_i^TyI + b_iI+ yW_i^T)(y - Wx).
\]
We will use this form to construct
a desired update rule with linear convergence. Let us consider an update step $g^s$ in expectation
over the code $x^*$ and and the noise $\eta$:
\begin{equation}
  \label{eq:7}
  g_i = -\E[\1_{x_i \neq 0}(W_i^TyI + b_iI + yW_i^T)(y - Wx)].
\end{equation}
To prove Theorem~\ref{thm:descent}, we compute $g_i$ according to the generative models described in~\eqref{eq:gen_model} and then argue the descent.
Here, we provide a proof sketch for (again) the simplest case of mixture-of-Gaussians; the full proof is deferred to Appendix~\ref{sec:dsc_proof}.


\proof[Proof of Theorem~\ref{thm:descent} (i)]

Based on Theorem~\ref{thm:consistency}, one can explicitly compute the expectation expressed in~\eqref{eq:7}. Specifically, the expected gradient $g_i$ is of the form:
\begin{align*}
  \label{eq:wm_gi}
  g_i &= -p_i\lambda_iA + p_i(\lambda_i^2 + 2b_i\lambda_i +  b_i^2)W_i + \gamma
\end{align*}
where $\lambda_i = \inprod{W^s_i}{A_i}$ and $\norm{\gamma} = O(n^{-w(1)})$. 
If we can find $b_i$ such that $\lambda_i^2 + 2b_i\lambda_i +  b_i^2 \approx \lambda_i$ for all $i$, $g_i$ roughly points in the same desired direction to $A_i$, and therefore, a descent property can be established via the following result:

\begin{Lemma}
  \label{lm:gmm_corr}
  Suppose $W$ is $\delta$-close to $A$ and the bias satisfies  $\abs{(b_i+\lambda_i)^2 - \lambda_i} \le 2(1-\lambda_i)$. Then:
  \begin{align*}
    2\inprod{g_i}{W_i - A_i} &\ge p_i(\lambda_i - 2\delta^2)\norm{W_i - A_i}^2 \\
    &+ \frac{1}{p_i\lambda_i}\norm{g_i}^2 - \frac{2}{p_i\lambda_i}\norm{\gamma}^2
  \end{align*}
\end{Lemma}
From Lemma~\ref{lm:gmm_corr}, one can easily prove the descent property
using~\cite{arora15_neural} (Theorem 6). We apply this lemma with learning
rate $\zeta = \max_i(1/p_i\lambda_i)$ and $\tau = \zeta p_i(\lambda_i - 2\delta^2) \in (0, 1)$ to achieve the descent as follows:
\[
  \norm{\widetilde{W}_i^{s+1} - A_i}^2 \leq (1 - \tau)\norm{W_i^{s} - A_i}^2 + O(n^{-K}),
\]
where $\widetilde{W}^{s+1} = W^s -\zeta g_s$ and $K$ is some constant greater than 1. Finally, we use
Lemma~\ref{lm:maintain_cls} to obtain the descent property for the normalized $W_i^{s+1}$. 

Now, we determine when the bias conditions in Theorem~\ref{thm:consistency} and Lemma~\ref{lm:gmm_corr} simultaneously hold for different choices of activation function. For the hard-thresholding function, since we do not need bias (i.e. $b_i = 0$ for every $i$), then ${\lambda_i(1-\lambda_i)} \le 2(1-\lambda_i)$ and this lemma clearly follows. 

On the other hand, if we encode $x = \relu(W^Ty + b)$, then we need every bias $b_i$ to satisfy $b_i \in [-1 + 2\delta^s\sqrt{k},\, -\delta^s]$ and $\abs{(b_i+\lambda_i)^2 - \lambda_i} \le 2(1-\lambda_i)$. Since $\lambda_i = \inprod{W^s_i}{A_i} \rightarrow 1$ and $\delta^s \rightarrow 0$, for the conditions of $b_i$ to hold, we require $b_i \rightarrow 0$ as $s$ increases. Hence, a fixed bias for the rectified linear unit would not work. Instead, we design a simple update for the bias (and this is enough to prove convergence in the ReLU case).

Here is our intuition. The gradient of $L$ with respect to $b_i$ is given by:
\begin{align*}
  \nabla_{b_i}L = -\sigma'(W_i^Ty + b_i)W_i^T(y - Wx)
\end{align*}
Similarly to the update for the weight matrix, we approximate this gradient with by replacing $\sigma'(W_i^Ty + b_i)$ with $\1_{x_i \neq 0}$, calculate the expected gradient and obtain:
\begin{align*}
  (g_b)_i &= -\E[W_i^T(y - Wx)\1_{x_i^* \neq 0}] + \gamma \\
          &= -\E[W_i^T(y- W_i(W_i^Ty + b_i)\1_{x_i^* \neq 0}] + \gamma \\
          &= -\E[(W_i^T- \norm{W_i}^2W_i^T)y + \norm{W_i}^2b_i\1_{x_i^* \neq 0}] + \gamma\\
          &= -p_ib_i + \gamma
\end{align*}
From the expected gradient formula, we design a very simple update for the bias:
$b^{s+1} = \sqrt{1 - \tau}b^{s}$ where $b^{0} = -1/\log n$, and show by 
induction that this choice of bias is sufficiently negative to make the consistency
result~\ref{thm:consistency} (ii) and (iii) hold at each step. 
At the first step, we have $\delta^0 \leq O^*(1/\log n)$, then
 \[b_i^{0} = -1/\log n \leq -\norm{W_i^{0} -
  A_i}.
  \] Now, assuming $b^s_i \le -\norm{W_i^s - A_i}$, we need to
prove that $b^{s+1} \le -\norm{W_i^{s+1} - A_i}$.

From the descent property at the step $s$, we have
\[
\norm{W_i^{s+1} - A_i} \le \sqrt{1-\tau}\norm{W_i^{s} - A_i} + o(\delta^s).
\] 

Therefore, $b_i^{s+1} = \sqrt{1-
\tau}b_i^{s} \leq -\sqrt{1-
\tau}\norm{W_i^{s} - A_i} \le -\norm{W_i^{s+1} - A_i} - o(\delta_s)$. As a result, $\abs{(b_i+\lambda_i)^2 - \lambda_i} \approx \lambda_i(1-\lambda_i) \le 2(1-\lambda_i)$. In addition, the condition of bias in the support consistency holds. By induction, we can guarantee the consistency at all the update steps. 
Lemma~\ref{lm:gmm_corr} and hence the descent results stated in (i.2) and
(iii.2) hold for the special case of the Gaussian mixture model.  \qedhere





\section{Experiments}

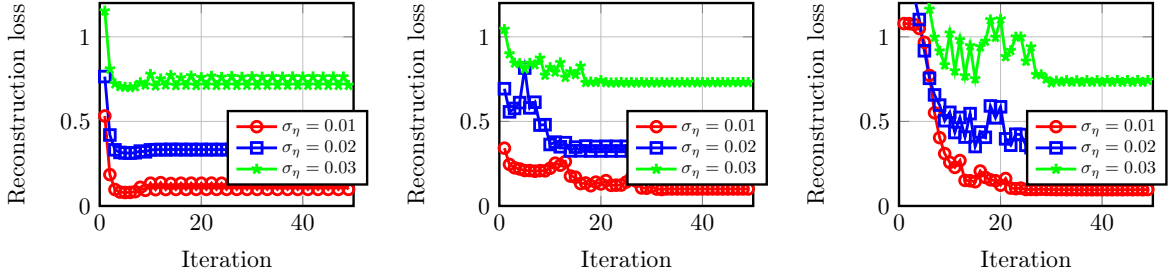
\begin{figure*}[!t]
\begin{center}
\begin{tabular}{ccc}
\begin{tikzpicture}[scale=.9]
\begin{axis}[
		width=3.75cm,
		height=3cm,
		scale only axis,
		xmin=0, xmax=50,
		xlabel = {Iteration},
		xmajorgrids,
		ymin=0, ymax=1.2,
		ylabel={Reconstruction loss},
		ymajorgrids,
		line width=1.0pt,
		mark size=1.5pt,
		legend style={nodes={scale=0.75, transform shape},at={(0.50,.1)},anchor=south west,draw=black,fill=white,align=left}
		]
\addplot  [color=red,
		solid, 
		very thick,
		mark=o,
		mark options={solid,scale=1.5},
		]
		table [x index = 0,y index=1]{results/reconstruction_loss.txt};
\addlegendentry{$\sigma_{\eta} = 0.01$}
\addplot [color=blue,
		solid, 
		very thick,
		mark=square,
		mark options={solid,scale=1.5},
		]
		table [x index = 0,y index=2]{results/reconstruction_loss.txt};
\addlegendentry{$\sigma_{\eta} = 0.02$}
\addplot  [color=green,
		solid, 
		very thick,
		mark=star,
		mark options={solid,scale=1.5},
		]
		table [x index = 0,y index=3]{results/reconstruction_loss.txt};
\addlegendentry{$\sigma_{\eta} = 0.03$}
\end{axis}
\end{tikzpicture}
&
\begin{tikzpicture}[scale=.9]
\begin{axis}[
		width=3.75cm,
		height=3cm,
		scale only axis,
		xmin=0, xmax=50,
		xlabel = {Iteration},
		xmajorgrids,
		ymin=0, ymax=1.2,
		ylabel={Reconstruction loss},
		ymajorgrids,
		line width=1.0pt,
		mark size=1.5pt,
		legend style={nodes={scale=0.75, transform shape},at={(0.50,.1)},anchor=south west,draw=black,fill=white,align=left}
		]
\addplot  [color=red,
		solid, 
		very thick,
		mark=o,
		mark options={solid,scale=1.5},
		]
		table [x index = 0,y index=1]{results/reconstruction_loss_pcab.5_4.txt};
\addlegendentry{$\sigma_{\eta} = 0.01$}
\addplot [color=blue,
		solid, 
		very thick,
		mark=square,
		mark options={solid,scale=1.5},
		]
		table [x index = 0,y index=2]{results/reconstruction_loss_pcab.5_4.txt};
\addlegendentry{$\sigma_{\eta} = 0.02$}
\addplot  [color=green,
		solid, 
		very thick,
		mark=star,
		mark options={solid,scale=1.5},
		]
		table [x index = 0,y index=3]{results/reconstruction_loss_pcab.5_4.txt};
\addlegendentry{$\sigma_{\eta} = 0.03$}
\end{axis}
\end{tikzpicture}
 &
\begin{tikzpicture}[scale=.9]
\begin{axis}[
		width=3.75cm,
		height=3cm,
		scale only axis,
		xmin=0, xmax=50,
		xlabel = {Iteration},
		xmajorgrids,
		ymin=0, ymax=1.2,
		ylabel={Reconstruction loss},
		ymajorgrids,
		line width=1.0pt,
		mark size=1.5pt,
		legend style={nodes={scale=0.75, transform shape},at={(0.50,.1)},anchor=south west,draw=black,fill=white,align=left}
		]
\addplot  [color=red,
		solid, 
		very thick,
		mark=o,
		mark options={solid,scale=1.5},
		]
		table [x index = 0,y index=1]{results/reconstruction_loss_randb.5_1.txt};
\addlegendentry{$\sigma_{\eta} = 0.01$}
\addplot [color=blue,
		solid, 
		very thick,
		mark=square,
		mark options={solid,scale=1.5},
		]
		table [x index = 0,y index=2]{results/reconstruction_loss_randb.5_1.txt};
\addlegendentry{$\sigma_{\eta} = 0.02$}
\addplot  [color=green,
		solid, 
		very thick,
		mark=star,
		mark options={solid,scale=1.5},
		]
		table [x index = 0,y index=3]{results/reconstruction_loss_randb.5_1.txt};
\addlegendentry{$\sigma_{\eta} = 0.03$}
\end{axis}
\end{tikzpicture}
\end{tabular}
\caption{\small\sl The learning curve  in training step using different initial estimate $W^0$. From left to right, the autoencoder is initialized by (i) some perturbation of the ground truth, (ii) PCA and (iii) random guess. \label{fig_exp_learning}}
\end{center}
\end{figure*}
\begin{figure}[!t]
\begin{center}
\resizebox{0.8\columnwidth}{!}{
\begin{tikzpicture}[scale=1.2]
\begin{axis}[
		width=6.5cm,
		height=4cm,
		scale only axis,
		xmin=0, xmax=50,
		xlabel = {Iteration},
		xmajorgrids,
		ymin=0, ymax=22,
		ylabel={$\norm{W - A}_F^2$},
		ymajorgrids,
		line width=1.0pt,
		mark size=1.5pt,
		legend style={nodes={scale=0.75, transform shape},at={(0.50,.5)},anchor=south west,draw=black,fill=white,align=left}
		]
\addplot  [color=red,
		solid, 
		very thick,
		mark=o,
		mark options={solid,scale=1.5},
		]
		table [x index = 0,y index=1]{results/matching_error.txt};
\addlegendentry{Init w/ perturbation}
\addplot [color=blue,
		solid, 
		very thick,
		mark=square,
		mark options={solid,scale=1.5},
		]
		table [x index = 0,y index=2]{results/matching_error.txt};
\addlegendentry{Init w/ PCA}
\addplot  [color=green,
		solid, 
		very thick,
		mark=star,
		mark options={solid,scale=1.5},
		]
		table [x index = 0,y index=3]{results/matching_error.txt};
\addlegendentry{Random init}
\end{axis}
\end{tikzpicture}
}
\caption{\small\sl Frobenius norm difference between the learned $W$ and the ground truth $A$ by three initialization schemes. \label{fig_sim_recovery}}
\end{center}
\end{figure}
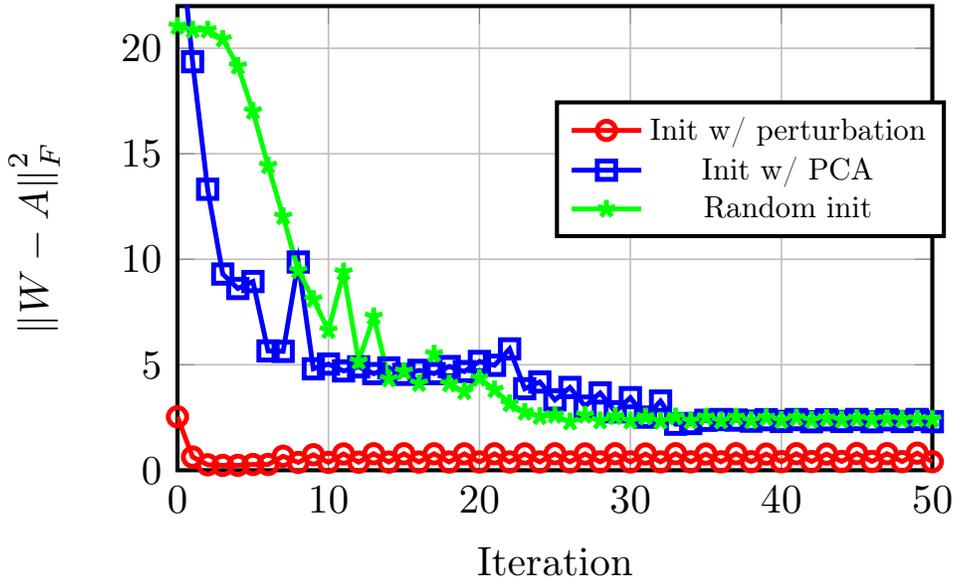

We support our theoretical results with some experiments on synthetic data sets under on the mixture-of-Gaussians model. We stress that these experimental results are \emph{not} intended to be exhaustive or of practical relevance, but rather only to confirm some aspects of our theoretical results, and shed light on where the theory falls short. 

We generate samples from a mixture of $m = 10$ Gaussians with dimension $n = 784$ using the model $y = Ax^* + \eta$. The means are the columns of $A$, randomly generated
according to $A_i \sim \mathcal{N}(0, \frac{1}{\sqrt{n}}I_n)$. To synthesize each sample $y$, we choose $x^*$ uniformly from the canonical bases $\{e_i\}_{i=1}^m$ and generate a Gaussian noise vector $\eta$ with independent entries and entry-wise standard deviation $\sigma_{\eta}$. We create a data set of $10,000$ samples in total for each Monte Carlo trial.

We consider a two-layer autoencoder with shared weights as described in Section~\ref{sec:autoencoder}, such that the hidden layer has $10$ units with $\relu$ activation. Then, we observe its gradient dynamics on the above data using three different initializations: (i) we initialize $W$ by adding small random perturbation to the groundtruth $A$ such that $W^{0} = A + \delta E$ for $\delta = 0.5$  with the perturbation $E \in \R^{784\times 10}$ generated according to $E_{ij} \sim \mathcal{N}(0, 1/\sqrt{n})$;
(ii) we perform principal component analysis of the data samples and choose the top $10$ singular vectors as $W^0$; (iii) we randomly generate $W$ with $W_{i} \sim \mathcal{N}(0, \frac{1}{\sqrt{n}}I_n)$.

For all three initializations, the bias $b$ of the encoder are initially set to $b =  -2.5\delta$. We train the weights $W$ with the batch gradient descent and update the bias using a fixed update rule $b^{s+1} = b^s/2$.

The learning rate for gradient descent is set fixed to $\zeta = m$. The number of descent steps is $T= 50$. We run the batch descent algorithm at each initialization with different levels of noise ($\sigma_\eta = 0.01, 0.02, 0.03$), then we observe the reconstruction loss over the data samples.


Figure~\ref{fig_exp_learning} shows the learning curve in the number of iterations. From the left, the first plot is the loss with the initial point $0.5$-close to $A$. The next two plots represent the learning using the PCA and random initializations. The gradient descent also converges when using the same step size and bias as described above. The convergence behavior is somewhat unexpected; \emph{even} with random initialization the reconstruction loss decreases to low levels when the noise parameter $\sigma_\eta$ is small. This suggests that the loss surface is perhaps amenable to optimization even for radius bigger than $O(\delta)$-away from the ground truth parameters, although our theory does not account for this.

In Figure~\ref{fig_sim_recovery} we show the Frobenius norm difference between the ground truth $A$ and final solution $W$ using three initialization schemes on a data set with noise $\sigma_\eta = 0.01$. Interestingly, despite the convergence, neither PCA nor random initialization leads to the recovery of the ground truth $A$. Note that since we can only estimate $W$ up to some column permutation, we use the Hungarian algorithm to compute matching between $W$ and $A$ and then calculate the norm.


\paragraph{Conclusions}

To our knowledge, the above analysis is the first to prove rigorous convergence of gradient dynamics for autoencoder architectures for a wide variety of (bilinear) generative models. Numerous avenues for future work remain --- finite sample complexity analysis; extension to more general architectures; and extension to richer classes of generative models.

\appendix

\section{Proof of Theorem~\ref{thm:consistency}}
\label{sec:cst_proof}

We start our proof with the following auxiliary claims.

\begin{Claim}
  \label{cl:bound_cross_prod}
  Suppose that $\max_i\norm{W_i-A_i}\le \delta$ and $\norm{W_i} = 1$.
  We have:
  \begin{enumerate}
  \item $\inprod{W_i}{\cAi} \geq 1 - \delta^2/2$ for any $i\in [m]$;
  \item $\abs{\inprod{W_i}{\cAj}} \leq \mu/\sqrt{n} + \delta$,
  for any $j \neq i \in [m]$;
  \item
  $\sum_{j \in S\backslash\{i\}}\inprod{W_i}{\cAj}^2 \leq O(\mu^2k/n + \delta^2)$
   for any $S \subset [m]$ of size at most $k$.
  \end{enumerate}
\end{Claim}

\proof The claims (i) and (ii) clearly follow from the $\delta$-closeness and $\mu$-incoherence properties as shown below.
\[
  \inprod{W_i}{\cAi} = 1 - (1/2)\norm{W_i - \cAi}^2 \geq 1
  - \delta^2/2,
\]
and
\[
  \abs{\inprod{W_i}{\cAj}} = \abs{\inprod{\cAi}{\cAj} +
    \inprod{W_i - \cAi}{\cAj}} \leq \mu/\sqrt{n} + \delta.
\]
For (iii), we apply Cauchy-Schwarz to bound each term inside the summation. Precisely, for any $j\neq i$,
\[
  \inprod{W_i}{\cAj}^2 \leq 2\bigl(\inprod{\cAi}{\cAj}^2 +
  \inprod{W_i - \cAi}{\cAj}^2\bigr) \leq 2\mu^2/n + 2\inprod{W_i
    - \cAi}{\cAj}^2.
\]
Together with $\norm{A} = O(\sqrt{m/n}) = O(1)$, we finish proving (iii) by noting that
\[
  \sum_{j \in S\backslash\{i\}}\inprod{W_i}{\cAj}^2 \leq  2\mu^2k/n
+ 2\norm{\cA{S}^{T}(W_i - \cAi)}_F^2 \leq 2\mu^2k/n +
2\norm{\cA{S}}^2\norm{W_i - \cAi}^2 \leq O(\mu^2k/n + \delta^2).
\]
\qedhere

\begin{Claim}
  \label{cl:bound_noise}
  Suppose $\norm{W_i} = 1$, then $\max_i \abs{\inprod{W_i}{\eta}} \leq \sigmae\log n$ holds with high probability.
\end{Claim}
\proof Since $\eta$ is a spherical Gaussian random vector and $\norm{W_i} = 1$,
$\inprod{W_i}{\eta}$ is Gaussian with mean $0$ and variance
$\sigmae^2$. Using the Gaussian tail bound for
$\inprod{W_i}{\eta}$ and taking the union bound
over $i = 1, 2, \dots, m$, we have that $\max_i\abs{\inprod{W_i}{\eta}} \leq
\sigmae\log n$ holds with high probability. \qedhere

\proof[Proof of Theorem~\ref{thm:consistency}] Denote $z = W^Ty + b$ and let $i\in [m]$ be fixed for a moment. (Later we use a union bound argument for account for all $i$).
Denote $S = \supp(x^*)$ and $R = S\backslash\{i\}$.
Notice that $x_i=0$ if $i\not\in S$ by definition. One can write the $i^{\textrm{th}}$ entry
$z_i$ of the weighted sum $z$ as
\begin{align*}
  z_i &= W_i^T(\cAS x_S^* + \eta) + b_i \\
  &=
\inprod{W_i}{\cAi}x_i^* + \sum_{j \in R}\inprod{W_i}{\cAj}x_j^*+ \inprod{W_i}{\eta} + b_i\\
  &= \inprod{W_i}{\cAi}x_i^* + Z_i + \inprod{W_i}{\eta} + b_i,
\end{align*}
where we write $Z_i = \sum_{j \in R}\inprod{W_i}{\cAj}x_j^*$.
Roughly speaking, since
$\inprod{W_i}{\cAi}$ is close to 1, $z_i$ approximately equals
$x_i^*$ if we can control the remaining terms. This will be made precise below separately for different generative models.






\subsection{Case (i): Sparse coding model}
\label{sec:consist_sc}

For this setting, the hidden code $x^*$ is $k$-sparse and is not restricted to non-negative values. The nonzero entries are mutually independent sub-Gaussian with mean $\mux=0$ and variance $\sigmax = 1$. Note further that $a_1 \in (0, 1]$ and $a_2 = \infty$ and the dictionary is incoherent and over-complete.

Since the true code takes both positive and negative values as well as sparse, it is natural to consider the hard thresholding activation.  The consistency is studied in~\citep{arora15_neural} for the case of sparse coding (see Appendix C and also work~\citep{nguyen18_double}, Lemma 8 for a treatment of the noise.)

\subsection{Case (ii) and (iii): Non-negative $k$-sparse model}
\label{sec:consist_nng}

Recall that $S = \supp(x^*)$ and that $x_j^* \in [a_1, a_2]$ for $j \in S$.
Cauchy-Schwarz inequality implies
\[
  \abs{Z_i} = \abs[\Big]{\sum_{j \in R}\inprod{W_i}{\cAj}x_j^*} \leq \sqrt{\sum_{j \in R}\inprod{W_i}{\cAj}^2}\norm{x^*}\leq a_2\sqrt{\frac{\mu^2k^2}{n} + k\delta^2},
\]
where we use bound (ii) in Claim~\ref{cl:bound_cross_prod} and
$\norm{x^*} \leq a_2\sqrt{k}$.

If $i \in S$, then \whp
\begin{align*}
  z_i &= \inprod{W_i}{\cAi}x_i^* + Z_i + \inprod{W_i}{\eta}
  \\
               &\geq (1 - \delta^2/2)a_1 - a_2\sqrt{\frac{\mu^2k^2}{n} +
                 k\delta^2} - \sigmae\log n +  b_i > 0
\end{align*}
for $b_i \geq -(1 - \delta)a_1 + a_2\delta\sqrt{k}$ and $a_2\delta\sqrt{k} \ll (1-\delta)a_1$, $k = O(1/\delta^2) = O(\log^2 n)$,  $\mu \leq \delta\sqrt{n}/k$, and $\sigmae = O(1/\sqrt{n})$.

On the other hand, when $i \notin S$ then \whp
\begin{align*}
  z_i &= Z_i + \inprod{W_i}{\eta} + b_i \\
  &\leq a_2\sqrt{\frac{\mu^2k^2}{n} +
                 k\delta^2} + \sigmae\log n + b_i \\
                 &\leq 0
\end{align*}
for $b_i \leq -a_2\sqrt{\frac{\mu^2k^2}{n} + k\delta^2} -
\sigmae\log n \approx -a_2\delta\sqrt{k}$. 

Due to the use of Claim 2, these results hold \whp\ uniformly for all $i$
and so $\supp(x) = S$ for $x = \relu(W^Ty+ b)$ \whp\ by We re-use the tail bound $\Prob[Z_i \ge \epsilon]$ given in~\citep{tran17}, Theorem 3.1.

Moreover, one can also see that  with high probability $z_i > a_1/2$ if $i \in S$ and $z_i < a_2\delta\sqrt{k} < a_1/4$ otherwise. This results hold \whp\ uniformly for all $i$ and so $x = \thres_{1/2}(z)$ has the same support as $x^*$ \whp
\qedhere

\section{Proof of Theorem~\ref{thm:descent}}
\label{sec:dsc_proof}

\subsection{Case (i): Mixture of Gaussians}
\label{sec:gmm}

We start with simplifying the form of $g_i$ using the generative model~\ref{thm:consistency} and Theorem~\ref{thm:consistency}. First, from the model we can have $p_i=\Prob[x_i^* \neq 0] = \Theta(1/m)$ and $\E[\eta] = 0$ and $\E[\eta\eta^T] = \sigmae^2I$. Second, by Theorem~\ref{thm:consistency} in (i), $\1_{x_i\neq 0} = x_i^* = 1$ with high probability. As such, under the event we have $x_i = \sigma(W_i^Ty + b_i) = (W_i^Ty + b_i)\1_{x_i^*\neq 0}$ for both choices of $\sigma$ (Theorem~\ref{thm:consistency}).

To analyze $g_i$, we observe that 
\[
  \gamma = \E[(W_i^TyI + b_iI + yW_i^T)(y - Wx)(\1_{x_i^* \neq 0} - \1_{x_i \neq 0})]
\]
has norm of order $O(n^{-w(1)})$ since the failure probability of the support consistency event is sufficiently small for large $n$, and the remaining term has bounded moments. One can write:
\begin{align*}
  g_i &= -\E[\1_{x_i^* \neq 0}(W_i^TyI + b_iI + yW_i^T)(y - Wx)] +
        \gamma \\
      &= -\E[\1_{x_i^* \neq 0}(W_i^TyI + yW_i^T + b_iI )(y - W_iW_i^Ty
        - b_iW_i)] + \gamma \\
      &= -\E[\1_{x_i^* \neq 0}(W_i^TyI
        \begin{aligned}[t]
        &+ yW_i^T)(I - W_iW_i^T)y] + b_i\E[\1_{x_i^* \neq 0}(W_i^TyI + yW_i^T)]W_i \\
        &- b_i\E[\1_{x_i^* \neq 0}(I - W_iW_i^T)y] + b_i^2 W_i\E[\1_{x_i^* \neq 0}] + \gamma
      \end{aligned}\\
      &= g_i^{(1)} + g_i^{(2)} + g_i^{(3)} +  p_ib_i^2W_i + \gamma,
\end{align*}
Next, we study each of $g_i^{(t)}$, $t=1,2,3$, by using the fact that $y = \cAi +
\eta$ as $x_i^* = 1$. To simplify the notation, denote $\lambda_i = \inprod{W_i}{\cAi}$.
Then
\begin{align*}
  g_i^{(1)} &= -\E[(W_i^T(\cAi + \eta)I + (\cAi + \eta)W_i^T)(I - W_iW_i^T)(\cAi + \eta)\1_{x_i^* \neq 0}] \\
      &= -\E[(\lambda_iI + \cAi W_i^T  + \inprod{W_i}{\eta}I + \eta W_i^T)(I - W_iW_i^T)(\cAi + \eta)\1_{x_i^* \neq 0}] \\
      &= -(\lambda_iI + \cAi W_i^T)(\cAi  - \lambda_iW_i)\Prob[x_i^* \neq 0] - \E[(\inprod{W_i}{\eta}I + \eta W_i^T)(I - W_iW_i^T)\eta\1_{x_i^* \neq 0}] \\
      &= -p_i\lambda_i\cAi  + p_i\lambda_i^2W_i - \E[(\inprod{W_i}{\eta}I + \eta W_i^T)(I - W_iW_i^T)\eta\1_{x_i^* \neq 0}],
\end{align*}
where we use $p_i=\Prob[x_i^* \neq 0]$ and denote $\norm{W_i} = 1$. Also, since $\eta$ is spherical Gaussian-distributed, we have:
\begin{align*}
  \E[(\inprod{W_i}{\eta}I + \eta W_i^T)(I - W_iW_i^T)\eta\1_{x_i^* \neq 0}] &= p_i\E[\inprod{W_i}{\eta}\eta - \inprod{W_i}{\eta}^2W_i] \\
                                                                                &= p_i\sigmae^2(1 - \norm{W_i}^2)W_i = 0,
\end{align*}
To sum up, we have
\begin{align}
  \label{eq:mg_g1}
  g_i^{(1)} &= -p_i\lambda_i\cAi  + p_i\lambda_i^2W_i
\end{align}
For the second term, 
\begin{align}
  \label{eq:mg_g2}
  g_i^{(2)} = b_i\E[\1_{x_i^* \neq 0}(W_i^TyI + yW_i^T)]W_i &= b_i\E[\1_{x_i^* \neq 0}(W_i^T(\cAi  + \eta)I + (\cAi  + \eta)W_i^T)]W_i \nonumber \\
                                                            &= b_i\E[(\lambda_iW_i + \norm{W_i}^2A_i )\1_{x_i^* \neq 0}] \nonumber \\
                                                            &= p_ib_i\lambda_iW_i
                                                              + p_ib_iA_i .
\end{align}
In the second step, we use the independence of spherical $\eta$ and $x$. Similarly, we can compute the third term:
\begin{align}
  \label{eq:mg_g3}
  g_i^{(3)} = - b_i(I - W_iW_i^T)\E[y\1_{x_i^* \neq 0}] &= -b_i(I - W_iW_i^T)\E[(\cAi  + \eta)\1_{x_i^* \neq 0}] \nonumber \\
                                                        &= -p_ib_i(I - W_iW_i^T)\cAi  \nonumber \\
                                                        &= -p_ib_i\cAi  + p_ib_i\lambda_iW_i
\end{align}

Putting~\eqref{eq:mg_g1},~\eqref{eq:mg_g2} and ~\eqref{eq:mg_g3} together, we have
\begin{align*}
  g_i 
      &= -p_i\lambda_i\cAi  + p_i(\lambda_i^2 + 2b_i\lambda_i +  b_i^2)W_i + \gamma
\end{align*}

Having established the closed-form for $g_i$, one can observe that when $b_i$ such that $\lambda_i^2 + 2b_i\lambda_i +  b_i^2 \approx \lambda_i$, $g_i$ roughly points in the same desired direction to $A^*$ and suggests the correlation of $g_i$ with $W_i - A_i $. Now, we prove this result.

\proof[Proof of Lemma~\ref{lm:gmm_corr}]
Denote $v = p_i(\lambda_i^2 + 2b_i\lambda_i +  b_i^2 - \lambda_i)W_i + \gamma$. Then
\begin{align}
  \label{eq:gi_v}
  g_i &= -p_i\lambda_i\cAi  + p_i(\lambda_i^2 + 2b_i\lambda_i +  b_i^2)W_i + \gamma \\
   \nonumber &= p_i\lambda_i(W_i - \cAi ) + v,
\end{align}
By expanding~\eqref{eq:gi_v}, we have
\[
2\langle v, W_i - A_i \rangle
=\frac{1}{p_i\lambda_i}\|g_i\|^2 - p_i\lambda_i\|W_i - A_i \|^2 - \frac{1}{p_i\lambda_i} \|v\|^2.
\]
Using this equality and taking inner product with $W_i - \cAi $ to both sides of~(\ref{eq:gi_v}),
we get  
\begin{align*}
  2\inprod{g_i}{W_i - A_i } =  p_i\lambda_i\norm{W_i - A_i }^2 + \frac{1}{p_i\lambda_i}\norm{g_i}^2 - \frac{1}{p_i\lambda_i}\norm{v}^2.
\end{align*}
We need an upper bound for $\norm{v}^2$. Since $$\abs{(b_i+\lambda_i)^2 - \lambda_i} \le 2(1-\lambda_i)$$ and $$2(1 - \lambda_i) = \norm{W_i - A_i }^2,$$ we have:
\[
  \abs{(b_i+\lambda_i)^2 - \lambda_i} \le \norm{W_i - A_i }^2 \le  \delta\norm{W_i - A_i }
\]
Notice that
\begin{align*}
  \norm{v}^2 &= \norm{p_i(\lambda_i^2 + 2b_i\lambda_i +  b_i^2 - \lambda_i)W_i + \gamma}^2  \\
  &\le 2p_i^2\delta^2\norm{W_i - A_i }^2 + 2\norm{\gamma}^2 .
\end{align*}
Now one can easily show that
\[
  2\inprod{g_i}{W_i - A_i } \ge p_i(\lambda_i - 2\delta^2)\norm{W_i - A_i }^2 + \frac{1}{p_i\lambda_i}\norm{g_i}^2 - \frac{2}{p_i\lambda_i}\norm{\gamma}^2.
\]
\qedhere

\subsection{Case (ii): General $k$-Sparse Coding}
\label{sec:sc}

For this case, we adopt the same analysis as used in Case 1. The difference lies in the distributional assumption of $x^*$, where nonzero entries are independent sub-Gaussian.
Specifically, given the support $S$ of size at most $k$ with $p_i = \Prob[i \in S] = \Theta(k/m)$ and $p_{ij} = \Prob[i, j \in S] = \Theta(k^2/m^2)$, we suppose $\E[x_i^*|S] = 0$ and $\E[x_S^{*}x_S^{*T}|S] = I$.
For simplicity, we choose to skip the noise, i.e.,~$y = A x^*$ for
this case. Our analysis is robust to iid additive Gaussian noise in the data; see~\citep{nguyen18_double} for a similar treatment.  Also, according to Theorem \ref{thm:consistency},
we set $b_i = 0$ to obtain support consistency.
With \emph{zero} bias, the expected update rule $g_i$ becomes
\[
  g_i = -\E[(W_i^Ty I + yW_i^T)(y - Wx)\1_{x_i \neq 0}].
\]
For  $S = \supp(x^*)$, then $y = \cAS x_S^*$.
Theorem~\ref{thm:consistency} in (ii) shows that $\supp(x) =  S$ \whp,
so under that event we can write $Wx = W_Sx_S =
W_S(W_S^Ty)$. Similar to the previous cases, $\gamma$ denotes a
general quantity whose norm is of order $n^{-w(1)}$ due to the
converging probability of the support consistency. Now, we substitute the forms of $y$ and $x$ into $g_i$: 
\begin{align*}
  g_i &= -\E[(W_i^TyI + yW_i^T)(y - Wx)\1_{x_i \neq 0}]\\
      &= -\E[(W_i^TyI + yW_i^T)(y - W_SW_S^Ty)\1_{x^*_i \neq 0}] + \gamma \\
      &= -\E[(I - W_SW_S^T)(W_i^T\cAS x_S^*)\cAS x_S^*\1_{x^*_i \neq 0}] - \E[(\cAS x_S^*)W_i^T(I - W_SW_S^T)\cAS x_S^*\1_{x^*_i \neq 0}] + \gamma \\
      &= g_i^{(1)} + g_i^{(2)} + \gamma.
\end{align*}
Write 
$$g_{i,S}^{(1)}=-\E[(I - W_SW_S^T)(W_i^T\cAS x_S^*)\cAS x_S^*\1_{x^*_i \neq 0}|S],$$
and 
$$g_{i,S}^{(2)} =- \E[(\cAS x_S^*)W_i^T(I - W_SW_S^T)\cAS x_S^*\1_{x^*_i \neq 0}|S],$$
so that $g_i^{(1)}=\E(g_{i,S}^{(1)})$ and $g_i^{(2)}=\E(g_{i,S}^{(2)})$.
It is easy to see that
$\E[x_j^*x_l^*\1_{x^*_i \neq 0}|S] =
1$ if $ i = j =
l\in S$ and $\E[x_i^*x_l^*\1_{x^*_i \neq 0}|S] = 0$ otherwise.
Therefore,
$g_{i,S}^{(1)}$ becomes
\begin{align}
  \label{eq:sc_g1}
  g_{i,S}^{(1)} &= -\E[(I - W_S W_S^T)(W_i^T\cAS x_S^*)\cAS x_S^*\1_{x^*_i \neq 0}|S] \\
  &= -\sum_{j, l \in S}\E[(I - W_S^T W_S)(W_i^TA_j )A_l x_j^*x_l^*\1_{x^*_i \neq 0}|S] \nonumber \\
                                                      &= -\lambda_i(I - W_S W_S^T)A_i ,
\end{align}
where we use the earlier notation $\lambda_i = W_i^TA_i $. Similar calculation of the
second term results in
\begin{align}
  \label{eq:sc_g2}
  g_{i,S}^{(2)} &= -\E[(\cAS x_S^*)W_i^T(I - W_S W_S^T)\cAS x_S^*\1_{x^*_i \neq 0}|S] \\
  &= -\E[\sum_{j \in S}x_j^*A_j W_i^T(I - W_S W_S^T)\sum_{l \in S}x_l^*A_l \1_{x^*_i \neq 0}|S] \nonumber \\
                                                                &= -\sum_{j, l \in S}\E[A_j W_i^T(I - W_S W_S^T)A_l x_j^*x_l^*\sgn(x_i^*)|S] \nonumber \\
                                                                &= -A_i W_i^T(I - W_S W_S^T)A_i 
\end{align}
Now we combine the results in~\eqref{eq:sc_g1} and~\eqref{eq:sc_g2} to
compute the expectation over $S$.
\begin{align}
  \label{eq:grad_3}
  g_i &= \E[g_{i,S}^{(1)}+ g_{i,S}^{(2)}] + \gamma\\
  &= -\E[\lambda_i(I - W_SW_S^T)\cAi  + \cAi W_i^T(I - W_S W_S^T)\cAi ] + \gamma \nonumber \\
      &= -\E[2\lambda_iA_i  - \lambda_i\sum_{j \in S}W_jW_j^TA_i   - \cAi W_i^T\sum_{j \in S}W_jW_j^TA_i ] + \gamma \nonumber \\
      &= -2p_i\lambda_iA_i  + \E[\lambda_i\sum_{j \in S}W_jW_j^T\cAi  + \sum_{j \in S}\inprod{W_i}{W_j}\inprod{A_i }{W_j}A_i  ] + \gamma \nonumber \\
      &= -2p_i\lambda_iA_i  + \E[\lambda_i^2W_i +  \sum_{j \in
        R}\inprod{A_i }{W_j}W_j + \lambda_i\norm{W_i}^2A_i  +
        \sum_{j \in R}\inprod{W_i}{W_j}\inprod{A_i}{W_j}\cAi ] +
        \gamma, \nonumber
\end{align}
where $p_i = \Prob[i \in S]$ and $R = S \backslash \{ i\}$. Moreover,
$\norm{W_i} = 1$, hence
\begin{align}
  g_i &= -p_i\lambda_iA_i + p_i\lambda_i^2W_i + \sum_{j \in [m]
        \backslash \{i\}}p_{ij}\lambda_i\inprod{A_i }{W_j}W_j + p_{ij}\inprod{W_i}{W_j}\inprod{A_i }{W_j}A_i ) + \gamma \nonumber \\
      &= -p_i\lambda_iA_i + p_i\lambda_i^2W_i +
        \lambda_iW_{-i}\diag(p_{ij})W_{-i}^T\cAi  + (W_i^TW_{-i}\diag(p_{ij})W_{-i}^TA_i )A_i  + \gamma,
\end{align}
for $W_{-i} = (W_1, \dots, W_{i-1}, W_{i+1}, \dots, W_m)$ with the $i^{\text{th}}$ column being removed, and $\diag(p_{ij})$ denotes the diagonal matrix formed by $p_{ij}$ with $j \in [m] \backslash\{i\}$.

Observe that ignoring lower order terms,
$g_i$ can be written as $p_i\lambda_i(W_i - A_i ) + p_i\lambda_i(\lambda_i - 1)W_i$, which roughly points in the same desired direction to $A$. Rigorously, we argue the following:
\begin{Lemma}
  \label{lm:sc_corr}
  Suppose $W$ is $(\delta, 2)$-near to $A$. Then
  \begin{equation*}
    2\inprod{g_i}{W_i - A_i } \ge p_i\lambda_i\norm{W_i - A_i }^2 + \frac{1}{p_i\lambda_i}\norm{g_i}^2 - O(p_ik^2/n^2\lambda_i)
  \end{equation*}
\end{Lemma}
\proof We proceed with similar steps as in the proof of Lemma~\ref{lm:gmm_corr}.
By nearness, $$\norm{W} \leq \norm{W - A} + \norm{A} \leq 3\norm{A} \leq O(\sqrt{m/n}).$$ Also, $p_i = \Theta(k/m)$ and $p_{ij} = \Theta(k^2/m^2)$. Then
\begin{align*}
  \norm{W_{-i}\diag(p_{ij})W_{-i}^T\cAi } &\leq p_i\norm{W_{-i}\diag(p_{ij}/p_i)W_{-i}^T} \\
  &\leq p_i\norm{W_{-i}}^2\max_{j \neq i} (p_{ij}/p_i) = O(p_ik/n).
\end{align*}
Similarly,
\[
  \norm{W_i^TW_{-i}\diag(p_{ij})W_{-i}^TA_i )A_i } \le O(p_ik/n).
\]

Now we denote 
$$v= p_i\lambda_i(\lambda_i - 1)W_i + \lambda_iW_{-i}\diag(p_{ij})W_{-i}^T\cAi  + (W_i^TW_{-i}\diag(p_{ij})W_{-i}^TA_i )A_i  + \gamma.$$ Then
\[
  g_i = p_i\lambda_i(W_i - A_i ) + v
\]
where $\norm{v} \leq p_i\lambda_i(\delta/2)\norm{W_i - A_i } + O(p_ik/n) + \norm{\gamma}$. Therefore, we obtain
\begin{equation*}
  2\inprod{g_i}{W_i - A_i } \ge p_i\lambda_i(1 - \frac{\delta^2}{2})\norm{W_i - A_i }^2 + \frac{1}{p_i\lambda_i}\norm{g_i}^2 - O(p_ik^2/n^2\lambda_i).
\end{equation*}
where we assume that $\norm{\gamma}$ is negligible when compared with $O(p_ik/n)$.
\qedhere

Adopting the same arguments in the proof of Case (i), we are able to get the descent property column-wise for the normalized gradient update with the step size $\zeta = \max_i(1/p_i\lambda_i)$ such that there is some $\tau \in (0, 1)$:
\[
  \norm{W_i^{s+1} - A_i }^2 \leq (1 - \tau)\norm{W_i^{s} - A_i }^2 + O(p_ik^2/n^2\lambda_i).
\]
Since $p_i = \Theta(k/m)$, Consequently, we will obtain the descent in Frobenius norm stated in Theorem~\ref{thm:descent}, item (ii).

\begin{Lemma}[Maintaining the nearness]
  $\norm{W - A} \leq 2\norm{A}.$
\end{Lemma}
\proof The proof follows from~\citep{arora15_neural} (Lemma 24 and Lemma 32).


\subsection{Case (iii): Non-negative $k$-Sparse Coding}

We proceed with the proof similarly to the above case of general $k$-sparse code. Additional effort is required due to the positive mean of nonzero coefficients in $x^*$. For $x = \sigma(W^Ty + b)$, we have the support recovery for both choices of $\sigma$ a shown in (ii) and (iii) of Theorem \ref{thm:consistency}. Hence we re-use the expansion in~\citep{tran17} to compute the expected approximate gradient.
Note that we standardize $W_i$ such that $\norm{W_i} = 1$ and ignore the noise $\eta$.

Let $i$ be fixed and consider the approximate gradient for the $i^{\text{th}}$ column of $W$. The expected approximate gradient has the following form:
\begin{equation*}
  g_i = -\E[\1_{x_i \neq 0}(W_i^TyI + b_iI + yW_i^T)(y - W x)] = \alpha_i W_i - \beta_i\cAi  + e_i,
\end{equation*}
where 
\begin{align*}
  \alpha_i = \sigmax p_i\lambda_i^2 + \sigmax \sum_{j \neq
  i}p_{ij}&\inprod{W_i}{A_j }^2 + 2\mux^2\sum_{j \neq
  i}p_{ij}\lambda_i\inprod{W_i}{A_j } + \mux^2\sum_{j \neq l \neq
      i}p_{ijl}\inprod{W_i}{\cAj }\inprod{W_i}{\cAl } \\
  &+ \begin{aligned}[t]
    2\mux p_ib_i\lambda_i + 2\mux\sum_{j \neq
      i}p_{ij}b_i\inprod{W_i}{\cAj } + p_ib_i^2;
  \end{aligned}
\end{align*}
\begin{align*}
  \beta_i = \sigmax p_i\lambda_i - \sigmax &\sum_{j \neq
  i}p_{ij}\inprod{W_i}{W_j}\inprod{A_i }{W_j} + \mux^2\sum_{j \neq
  i}p_{ij}\inprod{W_i}{\cAj } - \mux^2\sum_{j \neq   i}p_{ij}\inprod{W_i}{W_j}\inprod{W_j}{\cAj } \\
                                & -\begin{aligned}[t]
                                    \mux^2\sum_{j \neq l \neq i}p_{ijl}\inprod{W_i}{W_j}\inprod{W_j}{\cAl }
                                    - \mux\sum_{j\neq i}p_{ij}b_i\inprod{W_i}{W_j};
                                  \end{aligned}
\end{align*}
and $e_i$ is a term with norm $\norm{e_i} \leq O(\max{(\mux^2, \sigmax^2})p_ik/m)$ -- a rough bound obtained in~\citep{tran17} (see the proof of Lemma 5.2 in pages 26 and 35 of \citep{tran17}.) As a sanity check, by plugging in the parameters of the mixture of Gaussians to $\alpha_i, \beta_i$ and $e_i$, we get the same expression for $g_i$ in Case 1. We will show that only the first term in $\alpha_i$ is dominant except ones involving the bias $b_i$. The argument for $\beta_i$ follows similarly.

\begin{Claim}
  \label{cl:nnsc_alpha}
  \begin{align*}
  \alpha_i = \sigmax p_i\lambda_i^2 + \sigmax &O(p_ik/m) + 2\mux^2p_i\lambda_iO(k/\sqrt{m})
                                      + \mux^2O(p_ik^2/m) \\ &+ 2\mux p_ib_i\lambda_i + 2\mux p_{i}b_iO(k/\sqrt{m})+ p_ib_i^2.
\end{align*}
\end{Claim}
\proof We bound the corresponding terms in $\alpha_i$ one by one.
We start with the second term:
\begin{align*}
  \sum_{j \neq i}^mp_{ij}\inprod{W_i}{\cAj }^2 &\leq \max_{j \neq i}p_{ij}\sum_{j \neq i}^m\inprod{W_i}{A_j }^2 \\
  &\leq \max_{j \neq i}p_{ij}\norm{A^{T}_{-i}W_i}_F^2  \\
  &\leq O(p_ik/m), 
\end{align*}
since $p_{ij} = \Theta(k^2/m^2) = \Theta(p_ik/m)$. Similarly, we have
\begin{align*}
  \abs{\sum_{j \neq i}^mp_{ij}\inprod{W_i}{\cAj } } &= \abs{W_i^T\sum_{j \neq i}^mp_{ij}\cAj }  \\
  &\leq \norm{W_i}\norm{A}\sqrt{\sum_{j \neq i}p^2_{ij}} \\
  &\leq O(p_ik/\sqrt{m}),
\end{align*}
which leads to a bound on the third and the sixth terms.
Note that this bound will be re-used to bound the corresponding term in $\beta_i$.

The next term is bounded as follows:
\begin{align*}
  \sum_{\substack{j \neq l \\ j, l \neq i}}p_{ijl}\inprod{W_i}{\cAj }\inprod{W_i}{\cAl } &= W_i^T\sum_{\substack{j \neq l \\ j, l \neq i}}p_{ijl}\cAj \cAl^{T}W_i \\
  &\leq \norm[\Big]{\sum_{\substack{j \neq l \\ j, l \neq i}}p_{ijl}\cAj \cAl^{T}}\norm{W_i}^2 \\ 
  &\leq O(p_ik^2/m) ,
\end{align*}
where $M = \sum_{\substack{j \neq l \\ j, l \neq i}}p_{ijl}\cAj \cAl^{T} = A_{-i}QA^{T}_{-i}$ for $Q_{jl} = p_{ijl}$ for $j \neq l$ and $Q_{jl} = 0$ otherwise. Again, $A_{-i}$ denotes the matrix $W$ with its $i^{th}$ column removed. We have $p_{ijl} = \Theta(k^3/m^3) \le O(q_ik^2/m)$; therefore, $\norm{M} \leq \norm{Q}_F\norm{A}^2 \leq O(q_ik^2/m)$.
\qedhere

\begin{Claim}
  \label{cl:nnsc_beta}
  \begin{align*}
  \beta_i = \sigmax p_i\lambda_i - \sigmax &O(p_ik/m) + \mux^2O(p_ik/\sqrt{m}) - \mux^2O(p_ik/\sqrt{m}) \\
                                     &+ \mux^2 O(p_ik^2/m) - \mux b_iO(p_ik/\sqrt{m}).
\end{align*}
\end{Claim}

\proof We proceed similarly to the proof of Claim \ref{cl:nnsc_alpha}.
Due to nearness and the fact that $\|A^*\|=O(\sqrt{m/n}) = O(1)$,
we can conclude that $\norm{W} \leq O(1)$. For the second term, we have
\begin{align*}
  \norm{\sum_{j \neq i}p_{ij}\inprod{W_i}{W_j}\inprod{A_i }{W_j}} &= \norm{W_i^T\sum_{j \neq i}p_{ij}W_jW_j^TA_i } \\ 
  &\le \max_{j \neq i}p_{ij}\norm{W_{-i}W_{-i}^T}\norm{W_i}\norm{A_i } \\ 
  &\le O(p_ik/m),
\end{align*}
where $W_jW_j^T$ are p.s.d and so $0\preceq \sum_{j \neq i}p_{ij}W_jW_j^T \preceq
(\max_{j\neq i}p_{ij})(\sum_{j\neq i} W_j W_j^T) \preceq
\max_{j \neq i}p_{ij}W_{-i}W_{-i}^T$.
To bound the third one, we use the fact that $|\lambda_j| = |\inprod{W_j}{\cAj }| \le 1$.
Hence from the proof of Claim \ref{cl:nnsc_alpha},
\begin{align*}
  \norm{\sum_{j \neq   i}p_{ij}\inprod{W_i}{W_j}\inprod{W_j}{\cAj }} 
  &= \norm{\sum_{j \neq   i}p_{ij}\lambda_j\inprod{W_i}{W_j}} \\
  &\le \norm{W_i}\norm{W}\sqrt{\sum_{j \neq i}(p_{ij}\lambda_j)^2} \\ 
  &\leq O(p_ik/\sqrt{m}),
\end{align*}
which is also the bound for the last term. The remaining term can be bounded as follows:
\begin{align*}
  \norm{\sum_{j \neq l \neq i}p_{ijl}\inprod{W_i}{W_j}\inprod{W_j}{\cAl }} &\le \norm{\sum_{j \neq l \neq i}p_{ijl}W_jW_j^T\cAl } \\ 
  &\le  \sum_{l \neq i}\norm{p_{ijl}W_{-i}W_{-i}^T}\\
                                                                            &\leq \sum_{l \neq i}\max_{j \neq l \neq i}p_{ijl}\norm{W_{-i}}^2 \\ 
&\le O(p_ik^2/m).
\end{align*}
\qedhere

When $b_i = 0$, from~\eqref{cl:nnsc_alpha} and~\eqref{cl:nnsc_beta} and $b_i \in (-1, 0)$, we have:
\[
  \alpha_i = p_i(\sigmax \lambda_i^2 + 2\mux p_ib_i\lambda_i + b_i^2) + O(\max(\mux^2, \sigmax)k/\sqrt{m}) 
\]
and
\[
  \beta_i = \sigmax p_i\lambda_i + O(\max(\mux^2, \sigmax)k/\sqrt{m}),
\]
where we implicitly require that $k \leq O(\sqrt{n})$, which is even weaker than the condition $k = O(1/\delta^2)$ stated in Theorem~\ref{thm:consistency}. Now we recall the form of $g_i$:
\begin{equation}
  \label{eq:nnsc_gi}
  g_i = -\sigmax p_i\lambda_iA_i  + p_i(\sigmax \lambda_i^2 + 2\mux p_ib_i\lambda_i + b_i^2)W_i + v
\end{equation}
where $v = O(\max(\mux^2, \sigmax)k/\sqrt{m})A_i  + O(\max(\mux^2, \sigmax)k/\sqrt{m})W_i + e_i$.
Therefore $\norm{v} \leq O(\max(\mux^2, \sigmax)k/\sqrt{m})$.

\begin{Lemma}
  \label{lm:nnsc_corr}
  Suppose $A$ is $\delta$-close to $A^*$ and the bias satisfies $\abs{\sigmax \lambda_i^2 + 2\mux p_ib_i\lambda_i + b_i^2 - \sigmax\lambda_i} \leq 2\sigmax(1-\lambda_i)$, then
  \begin{equation*}
    2\inprod{g_i}{W_i - A_i } \ge \sigmax p_i(\lambda_i - 2\delta^2)\norm{W_i - A_i }^2 + \frac{1}{\sigmax p_i\lambda_i}\norm{g_i}^2 - O(\max(1, \sigmax/\mux^2) \frac{k^2}{p_i m})
  \end{equation*}
\end{Lemma}

The proof of this lemma and the descent is the same as that of Lemma~\ref{lm:gmm_corr} for the case of Gaussian mixture. Again, the condition for bias holds when $b_i = 0$ and the thresholding activation is used; but breaks down when the nonzero bias is set fixed across iterations.

Now, we give an analysis for a bias update. Similarly to the mixture of Gaussian case, the bias is updated as
\[
  b^{s+1} = b^{s}/C,
\]
for some $C > 1$. The proof remains the same to guarantee the consistency and also the descent.

The last step is to maintain the nearness for the new update. Since it is tedious to argue that for the complicated form of $g_i$, we can instead perform a projection on convex set $\mathcal{B} = \{W | \text{W is $\delta$-close to $A^*$ and } \norm{W} \leq 2\norm{A} \}$ to guarantee the nearness. The details can be found in~\citep{arora15_neural}.



\subsection{Auxiliary Lemma}

In our descent analysis, we assume a normalization for $W$'s columns after each descent update. The descent property is achieved for the unnormalized version and does not directly imply the $\delta$-closeness for that current estimate. In fact, this is shown by the following lemma:

\begin{Lemma}
  \label{lm:maintain_cls}
  Suppose that $\norm{W_i^s} = \norm{A_i } = 1$ and $\norm{W_i^s - A_i } \le \delta_s$. The gradient update $\widetilde{W}_i^{s+1}$ satisfies
  $
    \norm{\widetilde{W}_i^{s+1} - A_i } \le (1-\tau)\norm{W_i^s - A_i } + o(\delta_s)
  $.
  Then, for $\frac{1-\delta_s}{2-\delta_s} \le \tau < 1$, we have
  \[
    \norm{W_i^{s+1} - A_i } \leq (1 + o(1))\delta_s ,
  \]
  where ${W}_i^{s+1}=  \frac{\widetilde{W}_i^{s+1}}{\norm{\widetilde{W}_i^{s+1}}}$. 
\end{Lemma}
\proof Denote $w = \norm{\widetilde{W}_i^{s+1}}$. Using a triangle inequality and the descent property, we have 
\begin{align*}
  \norm{\widetilde{W}_i^{s+1} - wA_i } &= \norm{\widetilde{W}_i^{s+1} - A_i  + (1-w)A_i } \\
                            &\le \norm{\widetilde{W}_i^{s+1} - A_i } + \norm{(1-w)A_i } ~~~(\norm{A_i} = 1)\\ 
                            &\le (1-\tau)\norm{W_i^s - A_i } + (1-\tau)\norm{W_i^s - A_i }  + o(\delta_s)\\
                            &\le 2(1-\tau)\norm{W_i^s - A_i } + o(\delta_s). 
\end{align*}
At the third step, we use $\abs{1 - w} \leq \norm{\widetilde{W}_i^{s+1} - A_i } \le (1-\tau)\norm{W_i^s - A_i } + o(\delta_s)$. This also implies $w \geq 1 - (1-\tau - o(1))\delta_s$. Therefore,
\begin{align*}
  \norm{W_i^{s+1} - A_i } &\leq \frac{2(1-\tau)}{w}\norm{W_i^s - A_i } + o(\delta_s) \\
  &\le \frac{2(1-\tau)}{(1 + (1-\tau -o(1))\delta_s)}\norm{W_i^s - A_i } + o(\delta_s).
\end{align*}
This implies that when the condition $\frac{1+\delta_s}{2+\delta_s} \le \tau < 1$ holds, we get:
$$\norm{W_i^{s+1} - A_i } \leq (1 + o(1))\delta_s.$$
 \qedhere

\bibliographystyle{unsrtnat}
\bibliography{refs}

\end{document}